\documentclass[final]{cvpr}

\usepackage{times}
\usepackage{epsfig}
\usepackage{graphicx}
\usepackage{amsmath}
\usepackage{amssymb}
\usepackage{multirow}
\usepackage{color}
\usepackage[numbers, sort]{natbib}

\usepackage[pagebackref=true,breaklinks=true,colorlinks,bookmarks=false]{hyperref}

\def\red#1{\textbf{\textcolor[RGB]{244, 67, 54}{#1}}}
\def\blue#1{\textbf{\textcolor[RGB]{60, 120, 216}{#1}}}
\def\redl#1{\textcolor[RGB]{244, 67, 54}{#1}}
\def\bluel#1{\textcolor[RGB]{60, 120, 216}{#1}}

\def\cvprPaperID{2862} % *** Enter the CVPR Paper ID here
\def\confYear{CVPR 2021}

\begin{document}

%%%%%%%%% TITLE
\title{ReDet: A Rotation-equivariant Detector for Aerial Object Detection}

\author{Jiaming Han$^*$, 
Jian Ding$^*$, 
Nan Xue, 
Gui-Song Xia
\\[0.2cm]
Wuhan University, Wuhan, China \\
{\tt\small \{hanjiaming, jian.ding, xuenan, guisong.xia\}@whu.edu.cn}\\
}

\maketitle

\newcommand\blfootnote[1]{%
\begingroup
\renewcommand\thefootnote{}\footnote{#1}%
\addtocounter{footnote}{-1}%
\endgroup
}

\blfootnote{The study of this paper is funded by the National Natural Science Foundation of China (NSFC) under grant contracts No.61922065, No.61771350 and No.41820104006 and 61871299. It is also supported by Supercomputing Center of Wuhan University.} 
\blfootnote{$^*$Equal contribution.}
\blfootnote{Corresponding author: Gui-Song Xia (guisong.xia@whu.edu.cn).}

%%%%%%%%% ABSTRACT
\begin{abstract}
   Recently, object detection in aerial images has gained much attention in computer vision. Different from objects in natural images, aerial objects are often distributed with arbitrary orientation.
   Therefore, the detector requires more parameters to encode the orientation information, which are often highly redundant and inefficient.
   Moreover, as ordinary CNNs do not explicitly model the orientation variation, large amounts of rotation augmented data is needed to train an accurate object detector.
   In this paper, we propose a Rotation-equivariant Detector (ReDet) to address these issues, which explicitly encodes rotation equivariance and rotation invariance.
   More precisely, we incorporate rotation-equivariant networks into the detector to extract rotation-equivariant features, which can accurately predict the orientation and lead to a huge reduction of model size.
   Based on the rotation-equivariant features, we also present Rotation-invariant RoI Align (RiRoI Align), which adaptively extracts rotation-invariant features from equivariant features according to the orientation of RoI.
   Extensive experiments on several challenging aerial image datasets DOTA-v1.0, DOTA-v1.5 and HRSC2016, show that our method can achieve state-of-the-art performance on the task of aerial object detection.
   Compared with previous best results, our ReDet gains 1.2, 3.5 and 2.6 mAP on DOTA-v1.0, DOTA-v1.5 and HRSC2016 respectively while reducing the number of parameters by 60\% (313 Mb vs. 121 Mb). The code is available at: \url{https://github.com/csuhan/ReDet}.
\end{abstract}

\vspace{-3mm}
\section{Introduction}
This paper studies the problem of object detection in aerial images, a recently-emerged challenging problem in computer vision~\cite{xia2018dota}.
Different from objects in nature images, objects in aerial images are often distributed with arbitrary orientation.
To cope with these challenges, aerial object detection are usually formulated as an oriented object detection task by relying on Oriented Bounding Boxes (OBBs) representation instead of using Horizontal Bounding Boxes (HBBs)~\cite{xia2018dota,ding2018transformer,yang2019scrdet,yang2020arbitrary}.

\begin{figure}
\centering
      \includegraphics[width=\linewidth]{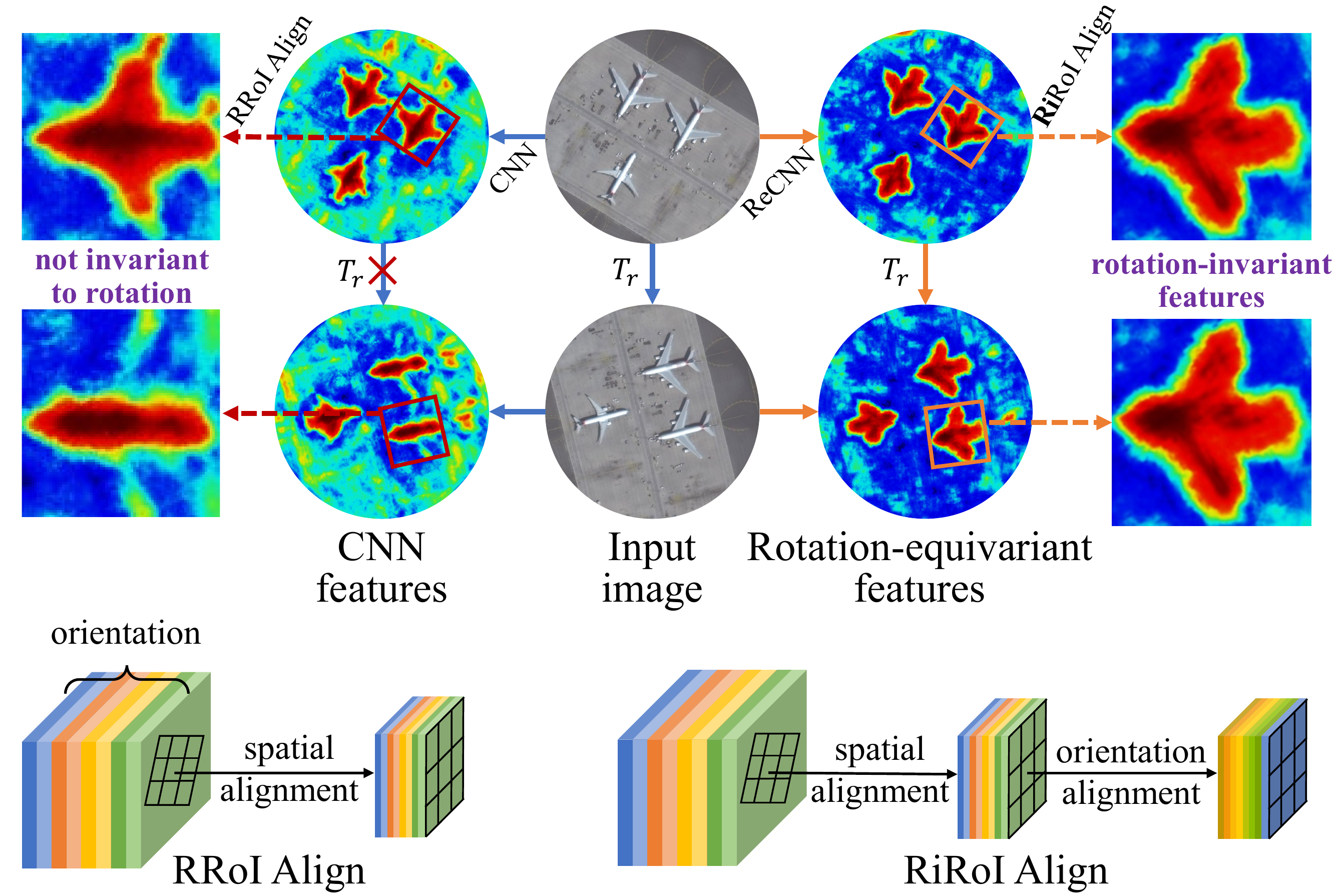} 
   % \vspace{-1mm}
   \caption{\textbf{Illustration of our method (top) and comparisons of RRoI warping (bottom)}. CNN features are not equivariant to the rotation $T_r$, \emph{i.e.}, feeding a rotated image to CNNs is not the same as rotating feature maps of the original image. Therefore, the corresponding RoI features are not invariant to rotation. In contrast, our method adopts rotation-equivariant CNNs (ReCNN) to extract rotation-equivariant features. Let $I$ and $\Phi$ be the input and ReCNN respectively, the equivariance of our method can be expressed as: $\Phi(T_rI)=T_r\Phi(I)$, \emph{i.e.}, applying a rotation $T_r$ to the image $I$ is the same as the rotation of features. Since we have obtained rotation-equivariant features, rotation-invariant features can be extracted by RRoI warping. While RRoI Align can only achieve rotation invariance in the spatial dimension, we present a novel Rotation-invariant RoI (RiRoI) Align to extract rotation-invariant features in both spatial and orientation dimensions.}
   \vspace{-3mm}
   \label{fig:rot_eq}
\end{figure}

% \textcolor{red}{
% Inspired by the success of Faster-RCNN~\cite{ren2017faster} on natural images, many well-designed two-stage oriented object detection approaches have reported promising results on challenging aerial image datasets~\cite{liu2017hrsc2016, xia2018dota}. 
% }
Recently, many well-designed oriented object detectors have been proposed and reported promising results on challenging aerial image datasets~\cite{liu2017hrsc2016,xia2018dota}.
In order to achieve accurate object detection in unconstrained aerial images, most of them are devoted to extract rotation-invariant features~\cite{ma2018arbitrary,ding2018transformer,yang2019r3det,han2020align}.
In practice, Rotated RoI (RRoI) warping (\emph{e.g.}, RRoI Pooling~\cite{ma2018arbitrary} and RRoI Align~\cite{ding2018transformer}) is the most commonly used method to extract rotation-invariant features, which can warp region features precisely according to the bounding boxes of RRoI in the 2D planar.
However, RRoI warping with regular CNN features can not produce exactly rotation-invariant features.
The rotation invariance is approximated by employing larger capacity networks and more training samples to model the rotation variation.
As shown in Fig.~\ref{fig:rot_eq}, the regular CNNs are not equivariant to the rotation, \emph{i.e.}, feeding a rotated image to CNNs is not the same as rotating feature maps of the original image. 
Therefore, region features warped from regular CNN feature maps are usually unstable and delicate as the orientation changes.

Some recently proposed methods~\cite{cohen2016gcnn,hoogeboom2018hexaconv,weiler2018learning} extend CNNs to larger groups and achieve rotation equivariance\footnote{Equivariance is a property that applying transformations to the input produces transformations of the feature in a predictable way.} with group convolutions~\cite{cohen2016gcnn}.
% In these rotation-equivariant networks, feature maps have additional orientation channels recording features from different orientations.
Feature maps of these methods have additional orientation channels recording features from different orientations.
However, directly applying the ordinary RRoI warping to rotation-equivariant features is unable to produce rotation-invariant features, as it can only warp region features in the 2D planar, \emph{i.e.}, the spatial dimension, while the orientation channels are still misaligned.
To extract completely rotation-invariant features, we also need to adjust the orientation dimension of feature maps according to the orientation of RRoI.

In this paper, we propose a Rotation-equivariant Detector (ReDet) to extract completely rotation-invariant features from rotation-equivariant features.
As shown in Fig.~\ref{fig:rot_eq}, our method consists of two parts: rotation-equivariant feature extraction and rotation-invariant feature extraction.
Firstly, we incorporate rotation-equivariant networks into the backbone to produce rotation-equivariant features, which can accurately predict the orientation and reduce the complexity of modeling orientation variations.
Since directly apply the RRoI warping still cannot extract rotation-invariant features from the rotation-equivariant features, we propose a novel Rotation-invariant RoI Align (RiRoI Align). It can warp region features according to the bounding boxes of RRoI in the spatial dimension and align features in the orientation dimension by circularly switching orientation channels and feature interpolation.
Finally, the combination of rotation-equivariant backbone and RiRoI Align forms our ReDet to extract completely rotation-invariant features for accurate aerial object detection.

\begin{figure}
   \begin{center}
      \includegraphics[width=\linewidth]{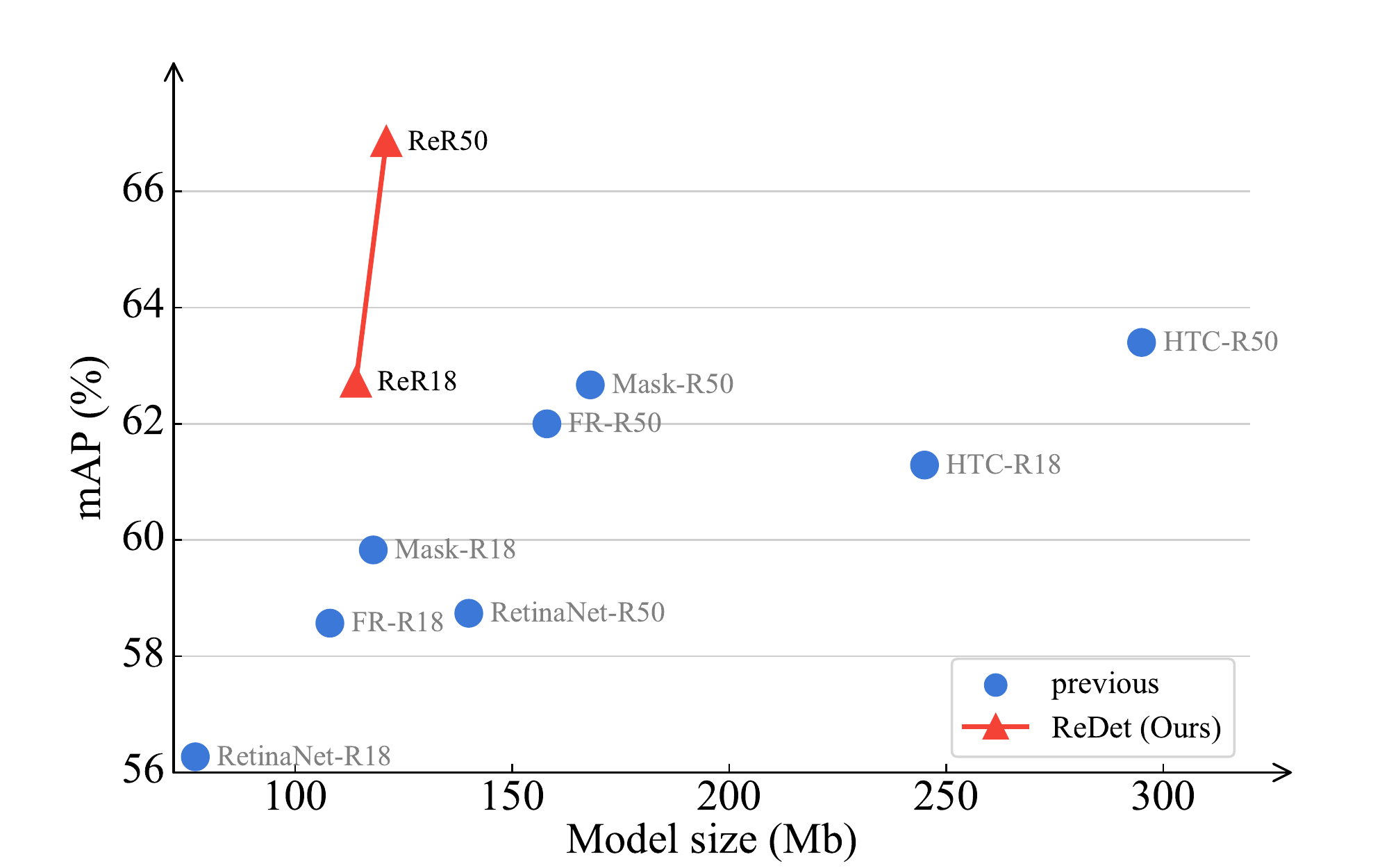}      
   \end{center}
   \vspace{-3mm}
   \caption{\textbf{Model size \emph{vs.} accuracy (mAP) on DOTA-v1.5.} We evaluate RetinaNet OBB~\cite{lin2017focal}, Faster R-CNN OBB (FR)~\cite{ren2017faster}, Mask R-CNN (Mask)~\cite{he2017maskrcnn} and Hybrid Task Cascade (HTC)~\cite{chen2019hybrid} with ResNet18 (R18) and ResNet50 (R50) backbones. Note all algorithms are our re-implemented version for DOTA, which is consistent with Tab.~\ref{tab:dota15_sota}. Our ReDet is tested with ReResNet18 (ReR18) and ReResNet50 (ReR50) backbones. Compared with other methods with R18/R50 backbones, our ReDet with a ReR18 backbone achieves competitive performance. Using a deeper backbone (ReR50), our ReDet outperforms all methods by a large margin and achieves better model size \emph{vs.} accuracy trade-off.}
   \vspace{-3mm}
   \label{fig:accuracy_parameter}
\end{figure}

Extensive experiments performed on the challenging aerial image datasets DOTA~\cite{xia2018dota} and HRSC2016~\cite{liu2017hrsc2016} demonstrate the effectiveness of our method.
We summary our contributions as:
\textbf{(1)} We propose a Rotation-equivariant Detector for high-quality aerial object detection, which encodes both rotation equivariance and rotation invariance. To our best knowledge, it is the first time that rotation equivariance has been systematically introduced into oriented object detection. 
\textbf{(2)} We design a novel RiRoI Align to extract rotation-invariant features from rotation-equivariant features. Different from other RRoI warping methods, RiRoI Align produces completely rotation-invariant features in both spatial and orientation dimensions.
\textbf{(3)} Our method achieves the state-of-the-art \textbf{80.10}, \textbf{76.80} and \textbf{90.46} mAP on DOTA-v1.0, DOTA-v1.5 and HRSC2016, respectively. Compared with previous best results, our method gains \textbf{1.2}, \textbf{3.5} and \textbf{2.6} mAP improvements. Compared with the baseline, our method shows consistent and substantial improvements and reduces the number of parameters by \textbf{60\%} (313 Mb \emph{vs.} 121 Mb). Moreover, our method achieves better model size \emph{vs.} accuracy trade-off (shown in Fig.~\ref{fig:accuracy_parameter}).

\vspace{-1mm}
\section{Related Works}
\subsection{Oriented Object Detection}
Unlike most general object detectors~\cite{girshick2014rich,girshick2015fast,ren2017faster,redmon2016you,liu2016ssd,lin2017focal,zhou2019objects} that use HBBs, oriented object detectors locate and classify objects with OBBs, which provide more accurate orientation information of objects. This is essential for detecting aerial objects with large aspect ratio, arbitrary orientation and dense distribution.
With the development of general object detection, many well-designed methods have been proposed for oriented object detection~\cite{xia2018dota,azimi2018towards,ding2018transformer,zhang2019cad,yang2019scrdet,pan2020dynamic,yang2020arbitrary}, showing promising performance on challenging datasets~\cite{xia2018dota,liu2017hrsc2016}.
To detect objects with arbitrary orientation, some methods~\cite{ma2018arbitrary, zhang2018toward, azimi2018towards} adopt numerous rotated anchors with different angles, scales and aspect ratios for better regression while increasing the computation complexity. Ding~\emph{et al.} proposed RoI Transformer~\cite{ding2018transformer} to transform Horizontal RoIs (HRoIs) into RRoIs, which avoids a large number of anchors.
Gliding vertex~\cite{xu2019gliding} and CenterMap~\cite{wang2020centermap} use quadrilateral and mask to accurately describe oriented objects, respectively. R$^3$Det and S$^2$A-Net align the feature between horizontal receptive fields and rotated anchors. DRN~\cite{pan2020dynamic} detects oriented objects with dynamic feature selection and refinement. CSL~\cite{yang2020arbitrary} regards angular prediction as a classification task to avoid discontinuous boundaries problem. Recently, some CenterNet~\cite{zhou2019objects}-based methods~\cite{pan2020dynamic,wei2020oriented,yi2020bbavector} show their advantages in detecting small objects.
The above methods are devoted to improving object representations or feature representations.
While our method is dedicated to improving the feature representation throughout the network: from the backbone to the detection head. Specifically, our method produces rotation-equivariant features in the backbone, significantly reducing the complexity in modeling orientation variations. In the detection head, the RiRoI Align extracts completely rotation-invariant features for robust object localization.

\begin{figure*}
\centering
      \includegraphics[width=.97\textwidth]{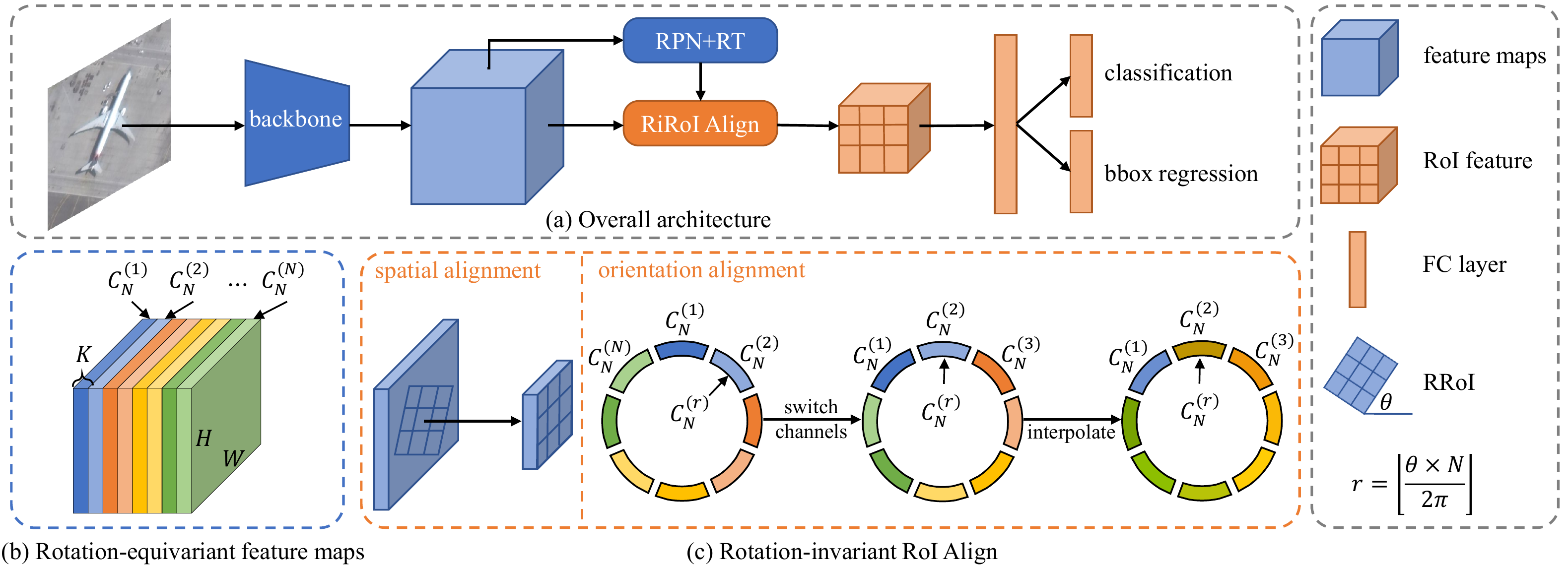}      
      \caption{\textbf{Overview of our proposed method}. \textbf{(a)} Overall architecture of the proposed Rotation-equivariant Detector. We first adopt the rotation-equivariant backbone to extract rotation-equivariant features, followed by an RPN and RoI Transformer (RT)~\cite{ding2018transformer} to generate RRoIs. Then we use a novel Rotation-invariant RoI Align (RiRoI Align) to produce rotation-invariant features for RoI-wise classification and bounding box (bbox) regression. \textbf{(b)} Rotation-equivariant feature maps. Under the cyclic group $C_N$, the rotation-equivariant feature maps with the size $(K,N,H,W)$ have $N$ orientation channels, and each orientation channel is corresponding to an element in $C_N$. \textbf{(c)} RiRoI Align. The proposed RiRoI Align consists of two parts: spatial alignment and orientation alignment. For an RRoI $(x,y,w,h,\theta)$, spatial alignment warps the RRoI from the spatial dimension, while orientation alignment circularly switches orientation channels and interpolates features to produce completely rotation-invariant features.}
   \vspace{-3mm}
   \label{fig:network}
\end{figure*}

\subsection{Rotation-equivariant Networks}
% Lenc \emph{et al.}~\cite{lenc2015understanding} show that AlexNet trained on imagenet spontaneously learns representations that are equivariant to flips, scaling and rotation, which supports the idea that equivariance is a good inductive bias for CNNs.
Cohen \emph{et al.} first proposed group convolutions~\cite{cohen2016gcnn} to incorporate 4-fold rotation equivariance into CNNs.
HexaConv~\cite{hoogeboom2018hexaconv} extends group convolutions to 6-fold rotation equivariance over hexagonal lattices.
To achieve rotation equivariance on more orientations, some methods~\cite{zhou2017orn,marcos2017rotation} resampling filters by interpolation, while other methods~\cite{worrall2017harmonic,weiler2018learning,weiler2019e2cnn} use harmonics as filters to produce equivariant features in the continuous domain.
The above methods gradually extend rotation equivariance to larger groups and achieve promising results on the classification task, while our method incorporates rotation-equivariant networks into the object detector, showing significant improvements on the detection task.
To our best knowledge, this is the first time that rotation equivariance has been systematically applied to oriented object detection.

\subsection{Rotation-invariant Object Detection}
The rotation-invariant feature is important for detecting arbitrary oriented objects.
However, CNNs show poor performance in modeling rotation variations, which means that more parameters are needed to encode the orientation information.
STN~\cite{jaderberg2015spatial} and DCN~\cite{dai2017deformable} explicitly model the rotation within the network and have been widely applied to oriented object detection~\cite{shi2016robust,ren2018deformable,ding2018transformer}.
Cheng \emph{et al.}~\cite{cheng2016ricnn} proposed a rotation-invariant layer that imposes an explicit regularization constraint to the objective.
Though the above methods can achieve approximated rotation invariance in the image-level, large amounts of training samples and parameters are needed. Besides, object detection requires instance-level rotation-invariant features.
Therefore, some methods~\cite{ma2018arbitrary,ding2018transformer} extend RoI warping~\cite{girshick2015fast} to RRoI warping, \emph{e.g.}, RoI Transformer~\cite{ding2018transformer} learns to transform HRoIs to RRoIs and warps region features with a rotated position sensitive RoI Align.
However, the regular CNNs are not rotation-equivariant. Therefore, even through the RRoI Align, we still cannot extract rotation-invariant features, as shown in Fig.~\ref{fig:rot_eq}.
Different from the aforementioned methods, our method proposes Rotation-invariant RoI Align (RiRoI Align) to extract rotation-invariant features from rotation-equivariant features.
Specifically, we incorporate rotation-equivariant networks into the backbone to produce rotation-equivariant features, then the RiRoI Align extracts completely rotation-invariant features from rotation-equivariant features in both spatial and orientation dimensions.

\section{Preliminaries}
\label{sec:rot_eq}
Equivariance is a property that applying transformations to the input produces transformations of the feature in a predictable way.
Formally, give a transformation group $G$ and a function $\Phi:X \rightarrow Y$, equivariance can be expressed as:
\begin{equation} \label{eq:equivariance}
   \Phi[T_g^X(x)]=T_g^Y[\Phi(x)] \quad \forall(x, g) \in(X, G),
\end{equation}
where $T_g$ indicates a group action in the corresponding space. 
Especially when $T_g^Y$ is identical for all $T_g^X$, equivariance becomes invariance. 

In common, CNNs are known to be translation equivariant. Let $T_t$ denotes an action of the translation group $(\mathbb{R}^2, +)$, and apply it to $K$-dimension feature maps $f:\mathbb{Z}^2 \rightarrow \mathbb{R}^{K}$, translation equivariance can be expressed as:
\begin{equation}\label{eq:trans_conv}
      \left[\left[T_t f\right] * \psi\right](x) = \left[T_t \left[f * \psi \right] \right](x),
\end{equation}
where $\psi: \mathbb{Z}^2 \rightarrow \mathbb{R}^{K}$ indicates the convolution filter and $*$ is the convolution operation.
Recently proposed methods~\cite{cohen2016gcnn,hoogeboom2018hexaconv,weiler2018learning} extend CNNs to large groups, achieving both translation and rotation equivariance.
Let $H$ denotes a rotation group, \emph{e.g.}, the cyclic group $C_N$ containing discrete rotations by angles multiple of $\frac{2\pi}{N}$.
We can define the group $G$ as the semidirect product of the translation group $(\mathbb{R}^2, +)$ and the rotation group $H$, \emph{i.e.}, $G \cong (\mathbb{R}^2, +) \rtimes H$.
By replacing $x \in (\mathbb{R}^2, +)$ with $g \in G$ in Eq.~\ref{eq:trans_conv}, the rotation-equivariant convolution can be defined as:
\begin{equation}\label{eq:rot_conv}
   \left[\left[T_g f\right] * \psi\right](g) = \left[T_g \left[f * \psi \right] \right](g).
\end{equation}

{\bf Rotation-equivariant Networks.} 
The regular CNNs consists of a series of convolution layers and enjoy the translation weight sharing.
Similarly, rotation-equivariant networks are a stack of rotation-equivariant layers with a higher degree of weight sharing, \emph{i.e.}, both translation and rotation.
Formally, let $\Phi=\{L_i|i\in \{1,2, \cdots, M\}\}$ denotes a network with $M$ rotation-equivariant layers under group $G$.
For a layer $L_i \in \Phi$, the rotation transformation $T_r$ can be preserved by the layer:
\begin{equation} \label{eq:rot_eq_layer}
   L_i[T_r(g)] = T_r[L_i(g)]\quad g\in G.
\end{equation}
If we apply $T_r$ to the input $I$ and feed it to the network $\Phi$, the transformation $T_r$\footnote{The transformation $T_r$ may have different formulations in different spaces, \emph{e.g.}, the input (image) space and the feature space. Here we do not distinguish it for simplicity. For a deeper discussion of rotation-equivariant networks, we refer the readers to~\cite{cohen2016gcnn} and~\cite{weiler2018learning}.} will be preserved by the whole network:
\begin{equation} \label{eq:rot_eq_layers}
   [\prod_{i=1}^M L_i](T_rI) = T_r[\prod_{i=1}^M L_i](I).
\end{equation}

{\bf Rotation-invariant Features.}
For any rotation transformations $T_r$ applied to the input, if its output remains unchanged, we say the output feature is rotation-invariant.
Rotation-invariant features can be divided into three levels: image-level, instance-level, and pixel-level. Here we mainly focus on the \textit{instance-level rotation-invariant feature}, which is more suitable for the object detection task.
Let $I_R \in I$ and $f_R \in f$ denotes an RoI of the image $I$ and feature maps $f$ ($f=\Phi(I)$), respectively.
Assume $I_R$ is a HRoI $(x,y,w,h)$ invariant to the orientation, where $(x,y)$, $w$ and $h$ denote the center point, width and height of the HRoI, respectively.
While $T_rI_R$ is an RRoI $(x,y,w,h,\theta)$ related to the orientation $\theta$.
Similar to Eq.~\ref{eq:rot_eq_layers}, for RoI $I_R$, the rotation equivariance can be expressed as:
\begin{equation} \label{eq:rot_eq_region}
   \Phi(T_rI_R) = T_r\Phi(I_R).
\end{equation}
If we regard HRoI $I_R$ as the rotation-invariant representation of RRoI $T_rI_R$ in the image $I$, $\Phi(I_R)$ can be regarded as the rotation-invariant representation of $\Phi(T_rI_R)$ in the corresponding feature space.
To get $\Phi(I_R)$, we need to know the rotation transformation $T_r$. Fortunately, $T_r$ is usually a function of the orientation $\theta$: $T_r=T(\theta)$.
% \begin{equation}\label{eq:rot_matrix}
%    T(\theta)=\begin{bmatrix}
%       cos(\theta)  & -sin(\theta) \\
%       sin(\theta)  & cos(\theta) \\
%       \end{bmatrix}.   
% \end{equation}
In practice, we can simply adopt a RRPN~\cite{ma2018arbitrary} or R-CNN to learn the orientation $\theta$ of an RRoI, as well as the transformation $T_r$.
Finally, the rotation-invariant feature $\Phi(I_R)$ can be obtained by applying an inverse transformation $T_r^{'}$ to Eq.~\ref{eq:rot_eq_region}:
\begin{equation} \label{eq:rot_eq_region_inv}
   \Phi(I_R) = T_r^{'}\Phi(T_rI_R).
\end{equation}

\section{Rotation-equivariant Detector}
\label{sec:methods}
This section presents details of the proposed Rotation-equivariant Detector (ReDet) to encode both rotation equivariance and rotation invariance.
First, we adopt rotation-equivariant networks as the backbone to extract rotation-equivariant features. As discussed before, directly applying the RRoI Align to rotation-equivariant feature maps cannot obtain the rotation-invariant features.
Therefore, we design a novel Rotation-invariant RoI Align (RiRoI Align), which produces RoI-wise rotation-invariant features from rotation-equivariant feature maps.
The overall architecture of ReDet is shown in Fig.~\ref{fig:network}.
For an input image, we feed it to the rotation-equivariant backbone. Then we adopt RPN to generate HRoIs, followed by an RoI Transformer (RT)~\cite{ding2018transformer} that transforms HRoIs to RRoIs. Finally, the RiRoI Align is adopted to extract rotation-invariant features for RoI-wise classification and bounding box regression.

\subsection{Rotation-equivariant Backbone}
Modern object detectors usually adopt deep CNNs as the backbone to automatically extract deep features with enriched semantic information, \emph{e.g.}, the widely used ResNet~\cite{he2016resnet} with Feature Pyramid Network (FPN)~\cite{lin2017feature}.
We also adopt ResNet with FPN as the baseline and implement a rotation-equivariant backbone, named Rotation-equivariant ResNet (ReResNet) with ReFPN.

Specifically, we re-implement all layers of the backbone with rotation-equivariant networks based on \texttt{e2cnn}~\cite{weiler2019e2cnn}, including convolution, pooling, normalization, non linearities, \emph{etc}.
Considering the computational budget, ReResNet and ReFPN are only equivariant to the discrete group $(\mathbb{R}^2, +) \rtimes C_N$, \emph{i.e.}, all translations and $N$ discrete rotations.
As is shown in Fig.~\ref{fig:network} (b), we can feed an image to the rotation-equivariant backbone to produce rotation-equivariant feature maps.
Unlike ordinary feature maps, the rotation-equivariant feature maps $f$ with the size $(K,N,H,W)$ have $N$ orientation channels: $f=\{f^{(i)}|i \in \{1,2,\cdots,N\}\}$, and feature maps of each orientation channel $f^{(i)}$ is corresponding to an element in $C_N$.

Compared with ordinary backbones, the rotation-equivariant backbone has the following advantages:
(a) \textbf{Higher degree of weight sharing.} As we have introduced that rotation-equivariant feature maps have an additional orientation dimension. Features from different orientations usually share the same filters with different rotation transformations, \emph{i.e.}, the rotation weight sharing.
(b) \textbf{Enriched orientation information.} For an input image with a fixed orientation, the rotation-equivariant backbone can produce features from multiple orientations. This is important for oriented object detection, which requires accurate orientation information.
(c) \textbf{Smaller model size.} Compared with the baseline, we have two choices when designing the backbone: similar computation or similar parameters. Typically, we keep similar computation with the baseline, \emph{i.e.}, preserving the same output channels. Due to the rotation weight sharing, our rotation-equivariant backbone shows a huge reduction of model size, about $1/N$ of parameters.

\subsection{Rotation-invariant RoI Align}
\label{sec:riroi_align}
As introduced in Sec.~\ref{sec:rot_eq}, for an RRoI $(x,y,w,h,\theta)$, we can extract rotation-invariant RoI features from rotation-equivariant feature maps with RRoI warping.
However, the ordinary RRoI warping can only align features in the spatial dimension, while the orientation dimension leaves misaligned.
Therefore, we propose RiRoI Align to extract completely rotation-invariant features.
As is shown in Fig.~\ref{fig:network} (c), RiRoI Align includes two parts: 
(a) \textbf{Spatial alignment.} For an RRoI $(x,y,w,h,\theta)$, spatial alignment warps it from feature maps $f$ to produce rotation-invariant region features $f_R$ in the spatial dimension, which is consistent with RRoI Align~\cite{ding2018transformer}.
(b) \textbf{Orientation alignment}. To ensure RRoIs with different orientations produce completely rotation-invariant features, we perform orientation alignment in the orientation dimension.
Specifically, for the output region features $\hat f_R$, we formulate orientation alignment as:
\vspace{-1mm}
\begin{equation}\label{eq:ori_align}
\hat f_R= Int(SC(f_R,r),\theta),\ r=\lfloor \theta N/2\pi \rfloor,
\end{equation}
where $SC$ and $Int$ denote the \textit{switching channels} and \textit{feature interpolation} operations, respectively.
For the region features $f_R$, we first calculate an index $r$, and circularly switch the orientation channels to make sure $C_N^{(r)}$ is the first orientation channel.
However, since the rotation equivariance is only achieved in the discrete group $C_N$, we also need to interpolate the feature if $\theta \notin C_N$.
More precisely, we interpolate the orientation feature with its nearest $l$ orientation channels. For example, the output feature of $i$-th orientation channel with $l=2$ can be expressed as:
\begin{equation}\label{eq:feature_int}
   \hat f_R^{(i)}= (1-\alpha) f_R^{(i)} + \alpha f_R^{(i+1)},
\end{equation}
where $\alpha=\theta N/2\pi-r$ indicates the distance factor for 1D-interpolation. Note that we use the \texttt{mod} function to ensure $i \in [1,N]$ (as well as $i+1$).

{\bf Comparison with RRoI Align+MaxPool.} Different from RiRoI Align, warping RoI features with RRoI Align and then maxpooling over the orientation dimension (\emph{i.e.}, orientation pooling) is another approach to extract rotation-invariant features.
The orientation pooling operation is usually adopted in classification tasks~\cite{cohen2016gcnn,zhou2017orn,weiler2018learning}. For each location in the feature map, it only preserves the orientation with the strongest response, while features from other orientations are abandoned.
However, we argue that the response from all orientations, no matter strong or weak, is indispensable for object recognition. In our RiRoI Align, features from all orientations are preserved and aligned with the orientation alignment operation.
We will conduct experiments to show the advantage of our RiRoI Align in Sec.~\ref{sec:exp}.

\section{Experiments and Analysis}
\label{sec:exp}

\subsection{Datasets}
{\bf DOTA~\cite{xia2018dota}} is the largest dataset for oriented object detection in aerial images with two released versions: {\bf DOTA-v1.0} and {\bf DOTA-v1.5}.
{\bf DOTA-v1.0} contains 2806 large aerial images with the size ranges from $800\times800$ to $4000\times4000$ and 188, 282 instances among 15 common categories: Plane (PL), Baseball diamond (BD), Bridge (BR), Ground track field (GTF), Small vehicle (SV), Large vehicle (LV), Ship (SH), Tennis court (TC), Basketball court (BC), Storage tank (ST), Soccer-ball field (SBF), Roundabout (RA), Harbor (HA), Swimming pool (SP), and Helicopter (HC).
{\bf DOTA-v1.5} is released for DOAI Challenge 2019\footnote[3]{\url{https://captain-whu.github.io/DOAI2019}} with a new category, Container Crane (CC) and more extremely small instances (less than 10 pixels). DOTA-v1.5 contains 402, 089 instances.
Compared with DOTA-v1.0, DOTA-v1.5 is more challenging but stable during training.

Following the settings in previous methods~\cite{ding2018transformer,han2020align}, we use both training and validation sets for training and the test set for testing. We crop the original images into $1024\times1024$ patches with a stride of 824. Random horizontal flipping is adopted to avoid over-fitting during training, and no other tricks are utilized.
For fair comparisons with other methods, we prepare multi-scale data at three scales \{0.5, 1.0, 1.5\}, and random rotation for training and testing.

{\bf HRSC2016~\cite{liu2017hrsc2016}} is a challenging ship detection dataset with OBB annotations, which contains 1061 aerial images with the size ranges from $300\times300$ to $1500\times900$. It includes 436, 181 and 444 images in the training, validation and test set, respectively. We use both training and validation sets for training and the test set for testing. All images are resized to (800, 512) without changing the aspect ratio. Random horizontal flipping is applied during training.

\subsection{Implementation Details}
{\bf ImageNet pretrain.} For the original ResNet~\cite{he2016resnet}, we directly use the ImageNet pretrained models from Pytorch~\cite{paszke2019pytorch}. 
For ReResNet, we implement it based on the \texttt{mmclassification}\footnote[4]{\url{https://github.com/open-mmlab/mmclassification}}. We train ReResNet on the ImageNet-1K with an initial learning rate of 0.1. All models are trained for 100 epochs and the learning rate is divided by 10 at \{30, 60, 90\} epochs. The batch size is set to 256. 

{\bf Fine-tuning on detection.} We adopt ResNet~\cite{he2016resnet} with FPN~\cite{lin2017feature} as the backbone of the baseline method. ReResNet with ReFPN is adopted as the backbone of our proposed ReDet.
For RPN, we set 15 anchors per location of each pyramid level. For R-CNN, we sample 512 RoIs with a 1:3 positive to negative ratio for training. For testing, we adopt 10000 RoIs (2000 for each pyramid level) before NMS and 2000 RoIs after NMS.
We adopt the same training schedules as \texttt{mmdetection}~\cite{chen2019mmdetection}. SGD optimizer is adopted with an initial learning rate of 0.01, and the learning rate is divided by 10 at each decay step. The momentum and weight decay are 0.9 and 0.0001, respectively. 
We train all models in 12 epochs for DOTA and 36 epochs for HRSC2016. We use 4 V100 GPUs with a total batch size of 8 for training and a single V100 GPU for inference.

\begin{table}
   \begin{center}
\small
\resizebox{\linewidth}{!}{%
    \begin{tabular}{c|c|c|c|c} \hline
        backbone    & group  & cls. (\%)      & det. (\%)  & size (Mb)   \\ \hline
        % R18-FPN     & -      & 70.17          & -                    & 53                 \\
        % ReR18-ReFPN & $C_8$  & 55.40          & 62.61                & \textbf{6}          \\
        R50-FPN     & -      & \textbf{76.55} & 65.03                & 103                 \\
        ReR50-ReFPN & $C_4$  & 72.81          & 65.43                & 24                 \\
        ReR50-ReFPN & $C_8$  & 71.20          & \textbf{66.86}       & 12                 \\
        ReR50-ReFPN &$C_{16}$& 61.60          & 64.36                & \textbf{6}        \\ \hline
    \end{tabular}
}
\end{center}
   \vspace{-3mm}
   \caption{\textbf{Performance comparisons of the rotation-equivariant backbone on classification (cls.) and detection (det.)}. group indicates the rotation group that the backbone is equivariant to. We report the top-1 accuracy on ILSVRC 2012 without FPN and the detection performance on DOTA-v1.5 test set in terms of mAP. The model size only includes the size of the backbone.}
   \vspace{-1mm}
   \label{tab:re_backbone}
\end{table}

\begin{table}
      \begin{center}
    \small
    % \resizebox{\linewidth}{!}{%/
        \begin{tabular}{c|c|c|c} \hline
            method                             & backbone    & mAP (\%)        & size (Mb)   \\ \hline
            \multirow{2}{*}{FR-O}  & R50-FPN     & 62.00           & 158         \\
                                               & ReR50-ReFPN & \textbf{62.36}  & \textbf{68} \\ \hline
            \multirow{2}{*}{RetinaNet-O}     & R50-FPN     & 58.74           & 140         \\
                                               & ReR50-ReFPN & \textbf{59.64}  & \textbf{34} \\ \hline
            % \multirow{2}{*}{S$^2$A-Net}        & R50-FPN     & 63.84           & 143         \\
            %                                    & ReR50-ReFPN & \textbf{65.82}  & \textbf{35} \\ \hline

        \end{tabular}
    % }
    \end{center}
      \vspace{-3mm}
      \caption{\textbf{The performance of rotation-equivariant backbone on other detectors}. Faster R-CNN OBB (FR-O) and RetinaNet OBB (RetinaNet-O) are our re-implemented version for OBBs.}
      \vspace{-1mm}
      \label{tab:other_method}
\end{table}

\begin{table}
      \begin{center}
    \small
    % \resizebox{\linewidth}{!}{%/
        \begin{tabular}{c|c|c} \hline
            method            & \#interpolate  & mAP (\%)               \\ \hline
            RRoI Align        & -              & 65.99                  \\
       RRoI Align+\emph{MP.}  & -              & 64.60 (-1.39)          \\
            RiRoI Align       & 1              & 66.44 (+0.45)          \\
            RiRoI Align       & 2              & \textbf{66.86 (+0.87)} \\
            RiRoI Align       & 4              & 66.32 (+0.33)          \\ \hline
        \end{tabular}
    % }
    \end{center}
      \vspace{-3mm}
      \caption{\textbf{Comparisons of our RiRoI Align with RRoI Align}. \#interpolate indicates the number of orientation channels used for interpolation (same as $l$ in Sec.~\ref{sec:riroi_align}). For an RRoI with the orientation $\theta$, we use its nearest \{1, 2, 4\} orientation channels to interpolate its features. \emph{MP.} is short for MaxPool. ReR50+ReFPN is adopted as the backbone.}
      \vspace{-1mm}
      \label{tab:riroi_align}
\end{table}

\begin{table}
   \begin{center}
    \small
    % \resizebox{\linewidth}{!}{%/
        \begin{tabular}{c|c|c|c|c} \hline
            method       & rot.     & schd.    & mAP (\%)       & training (h)\\ \hline
            % baseline     & $\times$ & 1x       & -            & -        \\
            ReDet        & $\times$ & 1x       & 62.62          & \textbf{8} \\
            baseline     &\checkmark& 1x       & 64.07          & 11         \\
            ReDet$^*$    & $\times$ & 1x       & 66.66          & 13         \\
            baseline     &\checkmark& 2x       & \textbf{67.34} & 22         \\  \hline
        \end{tabular}
    % }
    \end{center}
   \vspace{-3mm}
   \caption{\textbf{Comparison with rotation augmentation}. We compare the performance of the baseline method with rotation (rot.) augmentation and ReDet without rotation augmentation. ReDet$^*$ preserves a similar amount of parameters with the baseline. We report the mAP with R18 (for baseline) and ReR18 (for ReDet) backbone under the cyclic group $C_8$. For fair comparison, we randomly select rotation angles from $\{0,45,90,\cdots,315\}$.}
  \vspace{-1mm}
   \label{tab:rot_comp}
\end{table}

\begin{table}
   \begin{center}
    \resizebox{\linewidth}{!}{%/
    \begin{tabular}{c|c|c|c|c|c|c} \hline
        \multirow{2}{*}{method} & \multicolumn{3}{c|}{DOTA-v1.0}             & \multicolumn{3}{c}{HRSC2016}                   \\ \cline{2-7} 
                                & AP50     & AP75     & mAP                  & AP50     & AP75     & mAP                       \\ \hline
        baseline                & 75.62    & 48.37    & 46.13                & 90.18    & 80.48    & 68.17                     \\ 
        ReDet                   & 76.25    & 50.86    & \textbf{47.11(+0.98)}& 90.46    & 89.46    & \textbf{70.41(+2.24)}    \\ \hline
    \end{tabular}%
    }
    \end{center}

    % \begin{center}
    %     \resizebox{\textwidth}{!}{%/
    %     \begin{tabular}{c|c|c|c|c|c|c|c|c|c|c|c|c} \hline
    %         \multirow{2}{*}{method} & \multicolumn{3}{c|}{DOTA-v1.0}    & \multicolumn{3}{c|}{HRSC2016}     & \multicolumn{3}{c|}{MSRA-TD500}    & \multicolumn{3}{c}{ICDAR2015}     \\ \cline{2-13} 
    %                                 & AP50     & AP75     & mAP         & AP50     & AP75     & mAP         & P         & R        & F           & P        & R        & F           \\ \hline
    %         baseline                & 75.62    & 48.37    & 46.13       & 90.18    & 80.48    & 68.17       & 90.52     & 80.41    & 85.17       & 88.20    & 81.66    & 84.80       \\ 
    %         ReDet                   & 76.25    & 50.86    & \textbf{47.11(+0.98)}& 90.46    & 89.46    & \textbf{70.41(+2.24)}& 91.40     & 84.02    & \textbf{87.56(+2.39)}& 87.35    & 84.79    & \textbf{86.05(+1.25)}\\ \hline
    %     \end{tabular}%
    %     }
    %     \end{center}
   \vspace{-3mm}
   \caption{\textbf{Performance of the proposed ReDet on other datasets}. We report the performance on DOTA-v1.0 and HRSC2016 in COCO style. We use ReR50+ReFPN (\emph{resp.} R50+FPN) as the backbone of ReDet (\emph{resp.} baseline).}
  \vspace{-3mm}
   \label{tab:other_dataset}
\end{table}

\begin{table*}
   \begin{center}
\resizebox{\textwidth}{!}{%
\begin{tabular}{l|l|ccccccccccccccc|c}
\hline
method                                     & backbone    & PL           & BD           & BR           & GTF          & SV           & LV           & SH           & TC           & BC           & ST           & SBF          & RA           & HA           & SP           & HC           & mAP          \\ \hline 
\textbf{single-scale:}&&&&&&&&&&&&&&&& \\
FR-O~\cite{xia2018dota}                    & R101        & 79.42        & 77.13        & 17.70        & 64.05        & 35.30        & 38.02        & 37.16        & 89.41        & 69.64        & 59.28        & 50.30        & 52.91        & 47.89        & 47.40        & 46.30        & 54.13        \\
ICN~\cite{azimi2018towards}                & R101-FPN    & 81.36        & 74.30        & 47.70        & 70.32        & 64.89        & 67.82        & 69.98        & 90.76        & 79.06        & 78.20        & 53.64        & 62.90        & 67.02        & 64.17        & 50.23        & 68.16        \\
CADNet~\cite{zhang2019cad}                 & R101-FPN    & 87.80        & 82.40        & 49.40        & \blue{73.50} & 71.10        & 63.50        & 76.60        & \red{90.90}  & 79.20        & 73.30        & 48.40        & 60.90        & 62.00        & 67.00        & 62.20        & 69.90        \\
DRN~\cite{pan2020dynamic}                  & H-104       & 88.91        & 80.22        & 43.52        & 63.35        & 73.48        & 70.69        & 84.94        & 90.14        & 83.85        & 84.11        & 50.12        & 58.41        & 67.62        & 68.60        & 52.50        & 70.70        \\
CenterMap~\cite{wang2020centermap}         & R50-FPN     & 88.88        & 81.24        & \blue{53.15} & 60.65        & \red{78.62}  & 66.55        & 78.10        & 88.83        & 77.80        & 83.61        & 49.36        & \blue{66.19} & \blue{72.10} & \red{72.36}  & 58.70        & 71.74        \\
SCRDet~\cite{yang2019scrdet}               & R101-FPN    & \red{89.98}  & 80.65        & 52.09        & 68.36        & 68.36        & 60.32        & 72.41        & 90.85        & \red{87.94}  & \red{86.86}  & \red{65.02}  & \red{66.68}  & 66.25        & 68.24        & \blue{65.21} & 72.61        \\
R$^3$Det~\cite{yang2019r3det}              & R152-FPN    & \blue{89.49} & 81.17        & 50.53        & 66.10        & 70.92        & \blue{78.66} & 78.21        & 90.81        & 85.26        & 84.23        & \blue{61.81} & 63.77        & 68.16        & \blue{69.83} & \red{67.17}  & 73.74        \\
S$^2$A-Net~\cite{han2020align}             & R50-FPN     & 89.11        & \red{82.84}  & 48.37        & 71.11        & 78.11        & 78.39        & \blue{87.25} & 90.83        & 84.90        & 85.64        & 60.36        & 62.60        & 65.26        & 69.13        & 57.94        & \blue{74.12} \\
ReDet (Ours)                               & ReR50-ReFPN & 88.79        & \blue{82.64} & \red{53.97}  & \red{74.00}  & \blue{78.13} & \red{84.06}  & \red{88.04}  & \blue{90.89} & \blue{87.78} & \blue{85.75} & 61.76        & 60.39        & \red{75.96}  & 68.07        & 63.59        & \red{76.25}  \\ \hline
\textbf{multi-scale:}&&&&&&&&&&&&&&&& \\
RoI Trans.$^*$~\cite{ding2018transformer}  & R101-FPN    & 88.64        & 78.52        & 43.44        & 75.92        & 68.81        & 73.68        & 83.59        & 90.74        & 77.27        & 81.46        & 58.39        & 53.54        & 62.83        & 58.93        & 47.67        & 69.56        \\
O$^2$-DNet$^*$~\cite{wei2020oriented}      & H104        & 89.30        & 83.30        & 50.10        & 72.10        & 71.10        & 75.60        & 78.70        & \red{90.90}  & 79.90        & 82.90        & 60.20        & 60.00        & 64.60        & 68.90        & 65.70        & 72.80        \\
DRN$^*$~\cite{pan2020dynamic}              & H104        & 89.71        & 82.34        & 47.22        & 64.10        & 76.22        & 74.43        & 85.84        & 90.57        & 86.18        & 84.89        & 57.65        & 61.93        & 69.30        & 69.63        & 58.48        & 73.23        \\
Gliding Vertex$^*$~\cite{xu2019gliding}    & R101-FPN    & 89.64        & 85.00        & 52.26        & 77.34        & 73.01        & 73.14        & 86.82        & 90.74        & 79.02        & 86.81        & 59.55        & \red{70.91}  & 72.94        & 70.86        & 57.32        & 75.02        \\
BBAVectors$^*$~\cite{yi2020bbavector}      & R101        & 88.63        & 84.06        & 52.13        & 69.56        & 78.26        & 80.40        & 88.06        & 90.87        & \blue{87.23} & 86.39        & 56.11        & 65.62        & 67.10        & 72.08        & 63.96        & 75.36        \\
CenterMap$^*$~\cite{wang2020centermap}     & R101-FPN    & \blue{89.83} & 84.41        & 54.60        & 70.25        & 77.66        & 78.32        & 87.19        & 90.66        & 84.89        & 85.27        & 56.46        & \blue{69.23} & 74.13        & 71.56        & 66.06        & 76.03        \\
CSL$^*$~\cite{yang2020arbitrary}           & R152-FPN    & \red{90.25}  & \red{85.53}  & 54.64        & 75.31        & 70.44        & 73.51        & 77.62        & 90.84        & 86.15        & 86.69        & 69.60        & 68.04        & 73.83        & 71.10        & 68.93        & 76.17        \\
SCRDet++$^*$~\cite{yang2020scrdet++}       & R152-FPN    & 88.68        & \blue{85.22} & 54.70        & 73.71        & 71.92        & \blue{84.14} & 79.39        & 90.82        & 87.04        & 86.02        & 67.90        & 60.86        & 74.52        & 70.76        & \blue{72.66} & 76.56        \\
S$^2$A-Net$^*$~\cite{han2020align}         & R50-FPN     & 88.89        & 83.60        & \blue{57.74} & \red{81.95}  & \red{79.94}  & 83.19        & \red{89.11}  & 90.78        & 84.87        & \red{87.81}  & \red{70.30}  & 68.25        & \blue{78.30} & \blue{77.01} & 69.58        & \blue{79.42} \\
ReDet$^*$ (Ours)                           & ReR50-ReFPN & 88.81        & 82.48        & \red{60.83}  & \blue{80.82} & \blue{78.34} & \red{86.06}  & \blue{88.31} & \blue{90.87} & \red{88.77}  & \blue{87.03} & \blue{68.65} & 66.90        & \red{79.26}  & \red{79.71}  & \red{74.67}  & \red{80.10}  \\ \hline 
\end{tabular}%
}
\end{center}
   \vspace*{-3mm}
   \caption{\textbf{Comparisons with state-of-the-art methods on DOTA-v1.0 OBB Task}. H-104 means Hourglass 104. $^*$ indicates multi-scale training and testing. The results with \redl{red} and \bluel{blue} colors indicate the best and second-best results of each column, respectively.}
  \vspace*{-1mm}
   \label{tab:dota_sota}
\end{table*}

\begin{table*}
   \begin{center}
    \resizebox{\textwidth}{!}{%
    \begin{tabular}{l|cccccccccccccccc|c} \hline
    method          &PL& BD & BR & GTF & SV & LV & SH & TC & BC & ST & SBF & RA & HA & SP & HC & CC & mAP \\ \hline
    \textbf{OBB results:}&&&&&&&&&&&&&&&&& \\
    RetinaNet-O~\cite{lin2017focal}      &71.43&77.64&42.12&64.65&44.53&56.79&73.31&90.84&76.02&59.96&46.95&69.24&59.65&64.52&48.06&0.83&59.16 \\
    FR-O~\cite{ren2017faster}            &71.89&74.47&44.45&59.87&51.28&68.98&79.37&90.78&77.38&67.50&47.75&69.72&61.22&65.28&60.47&1.54&62.00 \\
    Mask R-CNN~\cite{he2017maskrcnn}     &76.84&73.51&49.90&57.80&51.31&71.34&79.75&90.46&74.21&66.07&46.21&70.61&63.07&64.46&57.81&9.42&62.67\\
    HTC~\cite{chen2019hybrid}            &77.80&73.67&51.40&63.99&51.54&73.31&80.31&90.48&75.12&67.34&48.51&70.63&64.84&64.48&55.87&5.15&63.40\\
    OWSR$^*$~\cite{li2019oswr}&-&-&-&-&-&-&-&-&-&-&-&-&-&-&-&-&74.90\\
    ReDet (Ours)                        &79.20&82.81&51.92&71.41&52.38&75.73&80.92&90.83&75.81&68.64&49.29&72.03&73.36&70.55&63.33&11.53&66.86 \\
    ReDet$^*$ (Ours)                    &88.51&86.45&61.23&81.20&67.60&83.65&90.00&90.86&84.30&75.33&71.49&72.06&78.32&74.73&76.10&46.98&\textbf{76.80} \\\hline
    \textbf{HBB results:}&&&&&&&&&&&&&&&&& \\
    RetinaNet-O~\cite{lin2017focal}     &71.66&77.22&48.71&65.16&49.48&69.64&79.21&90.84&77.21&61.03&47.30&68.69&67.22&74.48&46.16&5.78&62.49 \\
    FR-O~\cite{ren2017faster}           &71.91&71.60&50.58&61.95&51.99&71.05&80.16&90.78&77.16&67.66&47.93&69.35&69.51&74.40&60.33&5.17&63.85 \\
    HTC~\cite{chen2019hybrid}           &78.41&74.41&53.41&63.17&52.45&63.56&79.89&90.34&75.17&67.64&48.44&69.94&72.13&74.02&56.42&12.14&64.47\\
    Mask R-CNN~\cite{he2017maskrcnn}    &78.36&77.41&53.36&56.94&52.17&63.60&79.74&90.31&74.28&66.41&45.49&71.32&70.77&73.87&61.49&17.11&64.54\\
    OWSR$^*$~\cite{li2019oswr}&-&-&-&-&-&-&-&-&-&-&-&-&-&-&-&-&77.90\\
    ReDet (Ours)                        &79.51&82.63&53.81&69.82&52.76&75.64&87.82&90.83&75.81&68.78&49.11&71.65&75.57&75.17&58.29&15.36&67.66\\
    ReDet$^*$ (Ours)                    &88.68&86.57&61.93&81.20&73.71&83.59&90.06&90.86&84.30&75.56&71.55&71.86&83.93&80.38&75.62&49.55&\textbf{78.08}\\\hline
    \end{tabular}%
    }
\end{center}

   \vspace*{-3mm}
   \caption{\textbf{Performance comparisons on DOTA-v1.5 test set}. Note the results of Faster R-CNN OBB (FR-O)~\cite{ren2017faster}, RetinaNet OBB (RetinaNet-O)~\cite{lin2017focal}, Mask R-CNN~\cite{he2017maskrcnn} and Hybrid Task Cascade (HTC)~\cite{chen2019hybrid} are our re-implemented version for DOTA. OWSR~\cite{li2019oswr} is a method from DOAI 2019, and we report its single model performance for fair comparisons. The HBB results of our method are converted from OBB results by calculating the axis-aligned bounding boxes. $^{\mathbf{*}}$ means multi-scale training and testing.}
   \vspace*{-3mm}
   \label{tab:dota15_sota}
\end{table*}

\begin{table*}
   % \footnotesize
\small
\begin{center}
    % \resizebox{\textwidth}{!}{%
    \begin{tabular}{c|cccccccc|c} \hline
        method  & RC2~\cite{liu2017rrpnship} &RRPN~\cite{ma2018arbitrary}  & R$^2$PN~\cite{zhang2018toward} & RRD~\cite{liao2018rotation}  & RoI Trans.~\cite{ding2018transformer} & Gliding Vertex~\cite{xu2019gliding}  \\ 
        mAP     & 75.7                       &79.08                        & 79.6                 & 84.3                         & 86.2                                  & 88.2                                              \\ \hline
        method  & R$^3$Det~\cite{yang2019r3det} & DRN~\cite{pan2020dynamic} & CenterMap~\cite{wang2020centermap}      & CSL~\cite{yang2020arbitrary}      & S$^2$A-Net~\cite{han2020align}         &ReDet (Ours)       \\ 
        mAP     & 89.26                         & 92.7$^*$                  & 92.8$^*$                                & 89.62                             & 90.17 / 95.01$^*$  & \textbf{90.46} /  \textbf{97.63}$^*$ \\ \hline
        \end{tabular}
    % }    
\end{center}

  \vspace{-3mm}
  \caption{\textbf{Comparisons of state-of-the-art methods on HRSC2016}. $^*$ indicates that the result is evaluated under VOC2012 metrics, while other methods are all evaluated under VOC2007 metrics. We report both results for fair comparisons.}
   \vspace*{-1mm}
   \label{tab:hrsc2016_sota}
\end{table*}

\begin{figure*}
   \begin{center}
      \includegraphics[width=\linewidth]{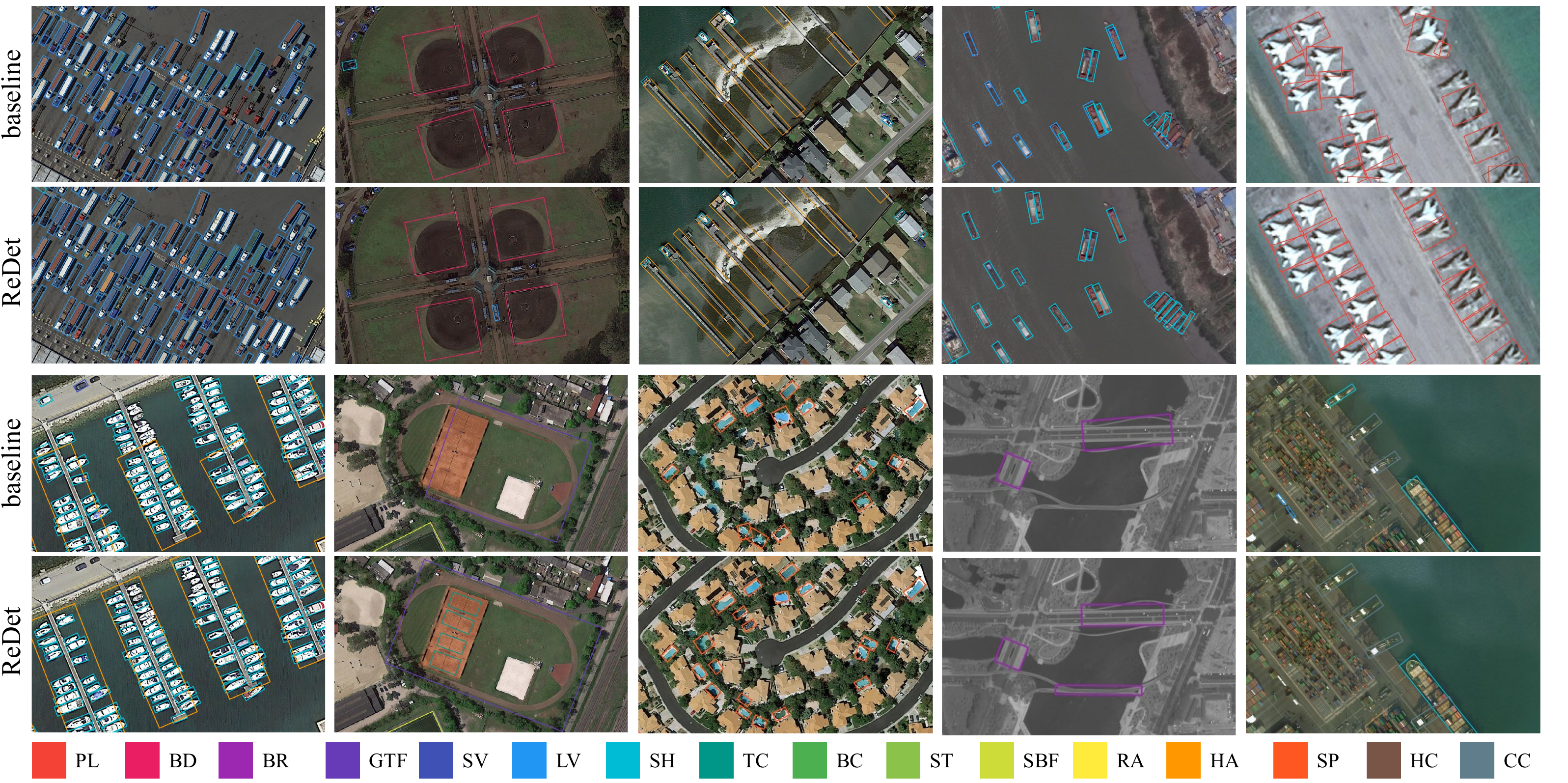}      
   \end{center}
   \vspace{-3mm}
   \caption{\textbf{Qualitative comparisons} between the proposed ReDet and the baseline method on DOTA-v1.5.}
  \vspace*{-3mm}
   \label{fig:dota_results}
\end{figure*}

\subsection{Ablation Studies}
In this section, we conduct a series of ablation experiments on DOTA-v1.5 test set to evaluate the effectiveness of our proposed method.
Note that we use the original ResNet+FPN and RRoI Align as the backbone and RoI warping method for the baseline method, respectively. 

{\bf Rotation-equivariant backbone.} We evaluate the effectiveness of rotation-equivariant backbone with ReResNet50+ReFPN under different settings. 
As shown in Tab.~\ref{tab:re_backbone}, compared to ResNet50, ReResNet50 achieves lower classification accuracy due to the reduction of parameters, but it obtains higher detection mAP.
We find the backbone under the cyclic group $C_8$ achieves better accuracy-parameter trade-off. ReResNet50+ReFPN under $C_8$ gains \textbf{1.83} detection mAP improvements with only {\bf 1/8} parameters (103 Mb \emph{vs.} 12 Mb).
Besides, we also extend ReResNet+ReFPN to other methods in Tab.~\ref{tab:other_method}. Both Faster R-CNN OBB and RetinaNet OBB with ReResNet50+ReFPN outperform its counterpart which further demonstrates the effectiveness of rotation-equivariant backbones.

{\bf Effectiveness of RiRoI Align.} As shown in Tab.~\ref{tab:riroi_align}, compared with RRoI Align, RiRoI Align shows significant improvements due to its orientation alignment mechanism.
While RRoI Align+MaxPool leads to a significant drop in mAP, indicating that the orientation pooling is undesirable in oriented object detection.
RiRoI Align with a $l=2$ interpolation achieves the highest \textbf{66.86} mAP and \textbf{0.87} mAP improvements than RRoI Align.
Besides, we find RiRoI Align with a $l=4$ interpolation only gains \textbf{0.33} mAP. The reason may be that too many interpolations hurt the equivariant property and inner relation between orientations.

{\bf Comparison with rotation augmentation.} From another perspective, our method can be viewed as a special in-network rotation augmentation, which learns from one orientation and can be applied to multiple orientations. In contrast, rotation augmentation enhances the network by generating samples with more orientations and usually requires more time to converge.
As shown in Tab.~\ref{tab:rot_comp}, although our method does not exceed the rotation augmented baseline under 1x schedule,
our ReDet$^*$, which preserves the similar amount of parameters, shows \textbf{2.59} mAP improvements with only \textbf{18\%} extra training time.
Moreover, the 2x baseline with rotation augmentation is \textbf{0.68} higher than our ReDet$^*$, but it takes \textbf{twice} the training time.

{\bf Performance on other datasets.} To prove the generalization of our proposed method, we also evaluate the performance of ReDet on DOTA-v1.0 and HRSC2016.
As is shown in Tab.~\ref{tab:other_dataset}, compared with the baseline, ReDet achieves better performance on both datasets.
Moreover, ReDet has significant improvements in \textbf{AP75} and \textbf{mAP}, which demonstrates its accurate localization capabilities.

\subsection{Comparisons with the State-of-the-Art}
{\bf Results on DOTA-v1.0.} As shown in Tab.~\ref{tab:dota_sota}, we compare our ReDet with other state-of-the-art methods on DOTA-v1.0 OBB Task. Without bells and whistles, our single-scale model achieves \textbf{76.25} mAP, outperforming all single-scale models and most multi-scale models.
With limited data augmentation (\emph{i.e.}, multi-scale data and random rotation), our method achieves state-of-the-art \textbf{80.10} mAP in the whole dataset, and obtains the best or second-best results among \textbf{12/15} categories.

{\bf Results on DOTA-v1.5.} Compared with DOTA-v1.0, DOTA-v1.5 contains many extremely small instances, which increases the difficulty of object detection. We report both OBB and HBB results on DOTA-v1.5 test set in Tab.~\ref{tab:dota15_sota}. With single-scale data, our method achieves \textbf{66.86} OBB mAP and \textbf{67.66} HBB mAP, outperforming RetinaNet OBB, Faster R-CNN OBB, Mask R-CNN~\cite{he2017maskrcnn} and HTC~\cite{chen2019hybrid} by a large margin.
Especially for the categories with small instances (\emph{e.g.}, HA, SP, CC) and large scale variations (\emph{e.g.}, PL, BD), our method performs better.
Besides, as shown in Fig.~\ref{fig:accuracy_parameter}, our ReDet achieves better parameter \emph{vs.} accuracy trade-off, which further demonstrates its efficiency.
Compared to previous best results by OWSR~\cite{li2019oswr}, our multi-scale model achieves state-of-the-art performance, about \textbf{76.80} OBB mAP and \textbf{78.08} HBB mAP. 
Qualitative comparisons between our ReDet and the baseline method are visualized in Fig.~\ref{fig:dota_results}.

{\bf Results on HRSC2016.} The HRSC2016 contains a lot of thin and long ship instances with arbitrary orientation.
We compare our ReDet with other state-of-the-art methods in Tab.~\ref{tab:hrsc2016_sota}. 
Our method achieves the state-of-the-art performance, \ie, with mAP of \textbf{90.46} and \textbf{97.63} under the VOC2007 and VOC2012 metrics, respectively.

\vspace{-1mm}
\section{Conclusions}
This paper presents a Rotation-equivariant Detector for aerial object detection, which consists of two parts: the rotation-equivariant backbone and the RiRoI Align.
The former produces rotation-equivariant features, while the latter extracts rotation-invariant features from rotation-equivariant features.
Extensive experiments on DOTA and HRSC2016 demonstrate the effectiveness of our method.
% show that our method can produce high-quality detection results and achieve state-of-the-art performance on challenging DOTA and HRSC2016.

{\small
\bibliographystyle{ieee_fullname}
\bibliography{egbib}
}

\end{document}